\documentclass{article}

\usepackage{arxiv}

\usepackage[utf8]{inputenc} % allow utf-8 input
\usepackage[T1]{fontenc}    % use 8-bit T1 fonts
\usepackage{hyperref}       % hyperlinks
\usepackage{url}            % simple URL typesetting
\usepackage{booktabs}       % professional-quality tables
\usepackage{amsfonts}       % blackboard math symbols
\usepackage{nicefrac}       % compact symbols for 1/2, etc.
\usepackage{microtype}      % microtypography
\usepackage{lipsum}
\usepackage{graphicx}
\graphicspath{ {./images/} }
\usepackage{paracol}

\title{Generating Statistical Charts with Validation-Driven LLM Workflows}

\author{
  Pavlin G. Poli\v{c}ar\thanks{Corresponding author: \texttt{pavlin.policar@fri.uni-lj.si}}\;\,\thanks{Equal contribution.} \\
    University of Ljubljana\\
    Faculty of Computer and Information Science\\
    Večna pot 113, 1000 Ljubljana, Slovenia\\
\And
  Andra\v{z} Pevcin\footnotemark[2] \\
    University of Ljubljana\\
    Faculty of Computer and Information Science\\
    Večna pot 113, 1000 Ljubljana, Slovenia\\
\And
    Bla\v{z} Zupan \\
    University of Ljubljana\\
    Faculty of Computer and Information Science\\
    Večna pot 113, 1000 Ljubljana, Slovenia\\
}
\date{}

\begin{document}

\maketitle

\begin{abstract}
Generating diverse, readable statistical charts from tabular data remains challenging for LLMs, as many failures become apparent after rendering and are not detectable from data or code alone. Existing chart datasets also rarely provide fully aligned artifacts, such as executable code, dataset context, and question-answer pairs. We present a structured LLM-based workflow that decomposes chart generation into dataset screening, plot proposal, code synthesis, rendering, validation-driven refinement, description generation, and question-answer generation. By incorporating rendered-output validation, the workflow addresses visualization-specific failure modes such as readability and semantic mismatch. It treats chart generation as an inspectable process rather than a one-shot prompt-to-code task, retaining each chart with its code, dataset context, description, and question-answer pairs. Applied to UCI datasets, the workflow produces 1,500 charts from 74 datasets, spanning 24 chart families and paired with 30,003 question-answer pairs. We evaluate 16 multimodal LLMs (MLLMs) on these chart-question pairs. The results show that chart-syntax questions are nearly saturated, while value extraction, comparison, and reasoning remain more challenging, illustrating the workflow's utility for diagnostic studies of chart-grounded multimodal reasoning.
\end{abstract}


\begin{figure}[t!]
  \centering
  \includegraphics[width=0.9\columnwidth]{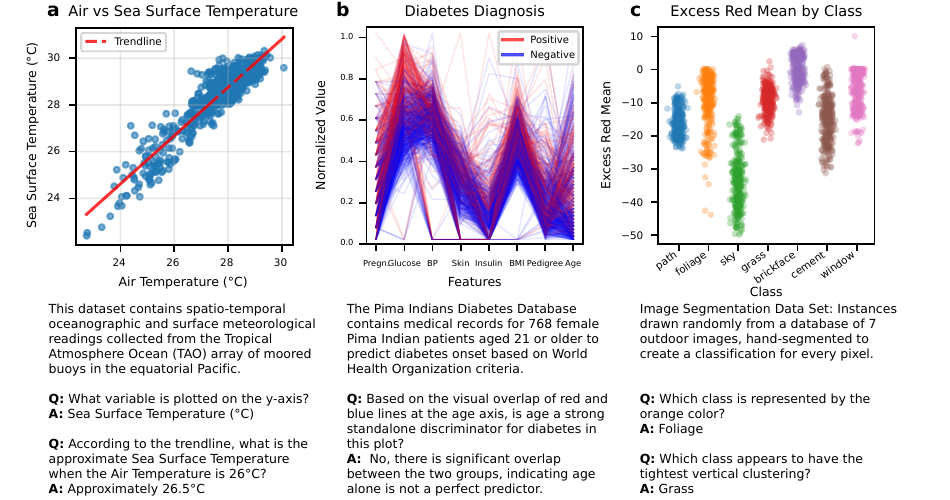}
  \caption{
    Three outputs generated by the proposed workflow. Each panel pairs a statistical chart generated from a public tabular dataset with the dataset context used by the workflow and selected chart-grounded question-answer pairs.
  }
  \label{fig:workflow-schematic}
\end{figure}

\section{Introduction}

Statistical charts combine visual structure, numerical content, and textual context, making them a central but challenging setting for multimodal reasoning. Answering chart-grounded questions requires models to interpret marks, axes, legends, scales, labels, and prompts while reasoning over encoded quantities. Numerous benchmarks show that this remains difficult even for strong chart-oriented systems and multimodal LLMs (MLLMs) \cite{Masry2022,Hoque2022,Islam2024,Masry2025,Huang2025}. Recent visualization research has increasingly examined how LLMs generate, interpret, and reason over visualizations, including work presented at previous IEEE VIS conferences \cite{Wang2026,Chen2025,Cui2025,Liu2025}. Useful chart outputs also extend beyond short answers to summaries, captions, open-ended explanations, and structured intermediate representations \cite{Kantharaj2022,Kantharaj2022a,Tang2023,Liu2025}, making charts a meaningful target for methods that generate charts together with descriptions and question-answer pairs.

Chart reasoning data are most useful when the chart image, dataset context, executable specification, description, and question-answer pairs are aligned, but existing chart collections rarely provide all of these outputs together. Curated collections from the web, scientific papers, or benchmark datasets are valuable but costly to construct, limited in coverage, and often tied to a single downstream task. Synthetic chart generation offers a complementary route, and recent work has used LLMs to synthesize chart data, plotting code, chart images, descriptions, and QA for model training or evaluation~\cite{Han2023,Yang2025,Huang2025}. However, these systems typically emphasize chart-understanding model training, benchmark construction, or constrained chart families rather than a workflow for generating statistical charts from public tabular datasets with rendered-output refinement and paired descriptions and typed question-answer pairs.

Generating statistical charts from raw tables is not a trivial one-shot code generation problem. A useful chart must identify meaningful variables and relationships, choose an appropriate visual encoding, generate executable plotting code, and produce a readable rendering. Visualization code synthesis remains difficult even in constrained settings~\cite{Chen2021}, and recent LLM-based systems show that important failures often appear only after rendering, making iterative debugging, rendered-output inspection, structured intermediate decisions, and explicit subtasks necessary~\cite{Yang2024,Chen2025,Xu2025,Tian2025}. This is especially important for charts, where readability, semantic fit, and the fidelity of later descriptions and questions depend on the rendered output rather than the source table or prompt alone. Recent work similarly argues for structured workflows with well-defined subtasks and checkpoints rather than open-ended prompting for complex visualization tasks~\cite{Wang2026}.

We present a structured LLM-based workflow for generating statistical charts from public tabular data. The workflow inspects the dataset, proposes suitable plots, generates plotting code, renders the chart, checks the rendered output for severe visual or semantic problems, and revises the chart when necessary. For each retained chart, the workflow preserves the intermediate decisions, final code, rendered versions, refinement feedback, description, and typed question-answer pairs. In the corpus generated for this paper, this choice space yielded 1,500 retained charts in which we observed 24 chart types.

Our contributions are threefold. First, we introduce a structured table-to-chart generation workflow that separates dataset screening, plot proposal, code generation, rendering, inspection, and refinement. We find that explicit rendered-output validation and a validation-driven refinement loop are important for producing semantically valid and readable charts, as many visualization failures are not detectable from code or data alone. Second, for each retained chart, we store the rendered image, executable plotting code, source-dataset context, generated description, typed question-answer pairs, and refinement feedback, resulting in a paired multimodal representation that supports inspection and downstream analysis. Third, we demonstrate one downstream use of these generated outputs by evaluating current MLLMs across chart families and question types, enabling fine-grained diagnostic analysis beyond aggregate accuracy.

\section{Related Work}

Prior work establishes charts as an important and difficult domain for multimodal reasoning. Benchmarks such as ChartQA~\cite{Masry2022}, PlotQA~\cite{Methani2020}, and related evaluations~\cite{Hoque2022,Islam2024,Masry2025,Wu2024,Xu2024,Huang2025} show that chart question answering requires coordinated visual, numerical, and linguistic reasoning, and that performance varies with chart type, realism, and question form. Beyond short-answer QA, chart summarization, captioning, open-ended QA, and compact chart representations show the value of faithful descriptions and explanatory responses~\cite{Kantharaj2022,Kantharaj2022a,Tang2023,Liu2025}. Chart-specific models and instruction datasets, including MATCHA~\cite{Liu2023}, UniChart~\cite{Masry2023}, ChartInstruct~\cite{Masry2024}, Omni-Chart-600K~\cite{Wang2025}, ChartLlama~\cite{Han2023}, EvoChart~\cite{Huang2025}, and recent synthetic training-data work~\cite{Yang2025}, further demonstrate the utility of chart-centered corpora for model training and evaluation. These resources primarily target chart understanding or instruction tuning; our workflow instead starts from public tabular datasets and generates the chart specification, executable code, chart rendering, dataset and chart descriptions, and typed question-answer pairs together.

Our work also builds on structured visualization generation. PlotCoder studies visualization code synthesis from notebook context~\cite{Chen2021}, while ChartGPT decomposes natural-language chart generation into staged data and visualization transformations~\cite{Tian2025}. MatPlotAgent generates and debugs plotting code, then uses an MLLM to inspect rendered drafts and propose revisions~\cite{Yang2024}; ChartIR similarly uses rendered-output comparison and iterative revision for chart-to-code reproduction~\cite{Xu2025}. VisEval shows that LLM-generated visualizations can fail in validity and readability, motivating checks beyond code execution~\cite{Chen2025}. Together with broader arguments for structured LLM workflows with explicit subtasks and checkpoints~\cite{Wang2026}, this prior work motivates our design: a table-to-chart workflow that renders and checks charts before generating descriptions and typed question-answer pairs for each retained output.

\section{End-to-End Statistical Chart Generation}

Our method is designed as a constrained multi-stage MLLM-based workflow. Each stage has a narrow responsibility, exposes an inspectable intermediate result, and passes structured information to the next stage. This structure keeps dataset screening, plot proposal, code synthesis, rendered-chart checking, refinement, description generation, and question labeling separate, so failures can be localized before a chart is retained. Separating rendering and post-render inspection exposes visualization failure modes--such as poor scaling, illegible labels, and misleading encodings--that are not detectable from the data schema or generated code alone. The design follows prior evidence that visualization-generation failures often become visible only after rendering, and that complex LLM-based visualization tasks benefit from explicit subtasks and checkpoints \cite{Yang2024,Chen2025,Wang2026}. Figure~\ref{fig:workflow-schematic} summarizes the pipeline.

\begin{figure}[t!]
  \centering
  \includegraphics[width=0.3\columnwidth]{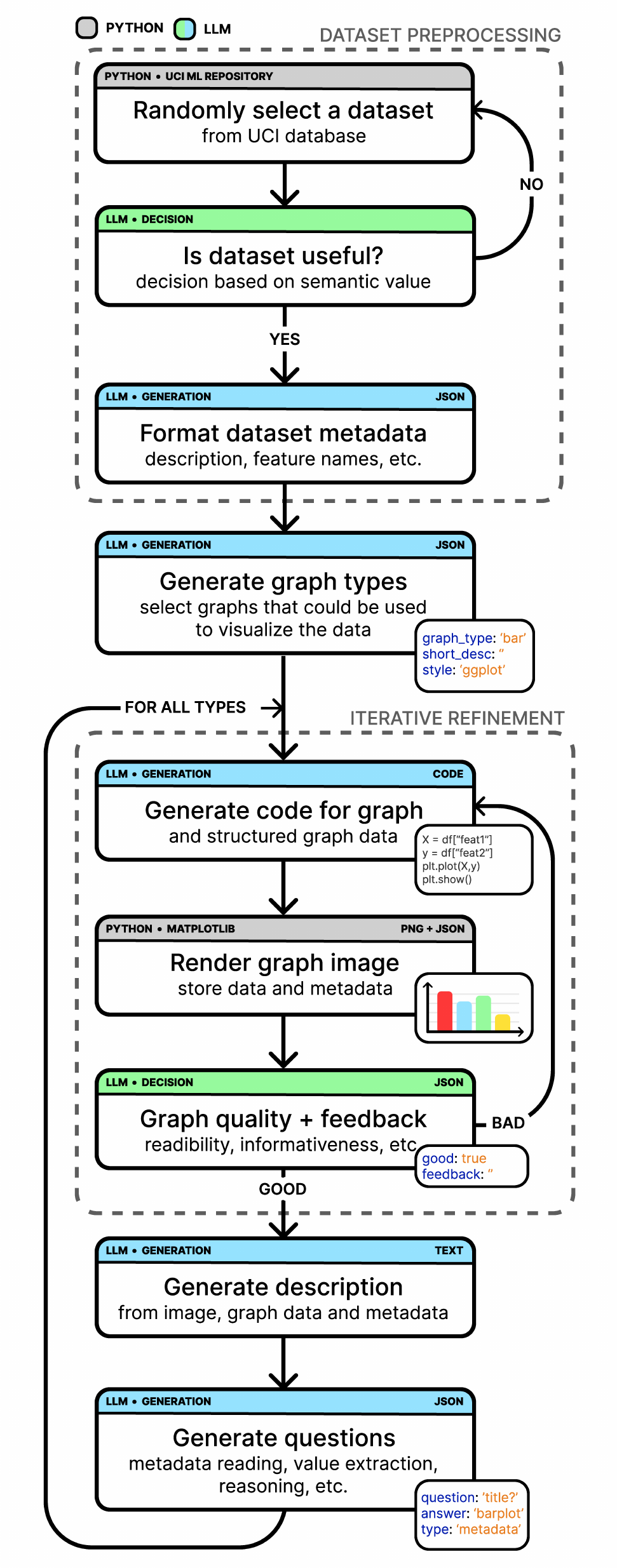}
  \caption{
    Schematic overview of the structured LLM-based generation workflow.
    The pipeline separates dataset screening, plot proposal, code generation, chart rendering, visual refinement, and description and QA generation.
    These stages keep intermediate decisions inspectable before a chart is retained.
  }
  \label{fig:workflow-schematic}
\end{figure}

\paragraph{Dataset selection and semantic preparation.}
We first identify candidate source datasets by randomly sampling tabular datasets from the UCI machine learning repository, retaining only those with at least 200 instances and at most 2,000 features. The first LLM stage then decides whether a sampled dataset is likely to support useful statistical charts. This screening step uses the dataset metadata, feature schema, and dataset description to reject datasets whose variables are mostly identifiers, opaque codes, poorly described fields, or otherwise unlikely to support meaningful comparisons, distributions, or relationships. For retained datasets, the same call rewrites the raw dataset description into a shorter, data-focused summary. A second structured call then rewrites feature names into clearer axis and legend labels when the dataset description supports a confident semantic interpretation.

\paragraph{Plot proposal.}
Given the cleaned dataset description, feature schema, and a small table preview, the next stage proposes ten candidate charts. Each proposal specifies a chart type, selected features, and a short description of what the chart should show. Generating chart proposals jointly instead of one-by-one encourages diversity in the chart types and selected features. The proposals are constrained to chart types that can be implemented with a standard scientific Python plotting stack.

\paragraph{Rendering and validation-aware refinement.}
Each candidate specification is rendered independently by a code-generation stage that produces executable Python using only the selected features. Besides saving the chart image, the generated code records the variables, encodings, transformations, filters, aggregations, and row counts used in the chart. These outputs ground later description and question generation in the chart that was actually rendered. After execution, an MLLM-based checking stage reviews the rendered chart and code for severe readability and semantic-fit problems, including clutter, label collisions, weak legends, poor scaling, indistinguishable groups, unreadable text, and encoding mismatches. If the checker reports a problem severe enough to warrant correction, the workflow sends the previous code and feedback back to the code-generation stage, which regenerates the same chart type while preserving the selected data variables whenever possible. In practice, many initially generated charts exhibit issues such as overplotting, unreadable axes, or ambiguous encodings, which are often improved through this refinement loop. This render-check-regenerate loop continues until the checker indicates that remaining issues do not warrant regeneration, or until three visual-correction retries have been attempted. Chart candidates for which the checker still requests correction after the third retry are discarded before description and question-answer generation.

\paragraph{Description and QA generation.}
For each retained chart, the final stage generates a chart description based on the chart, the plotting code, recorded chart details, and dataset context. It then targets 20 chart-grounded question-answer pairs per chart, with a difficulty mix of 7 easy, 6 medium, and 7 hard questions. Questions must be answerable from the chart and provided context rather than external knowledge. A final labeling step assigns each question one type: \textit{chart syntax}, \textit{value extraction}, \textit{comparison}, \textit{trends}, or \textit{reasoning}.

\section{Downstream Diagnostic Evaluation}

\begin{figure*}[t!]
  \centering
  \includegraphics{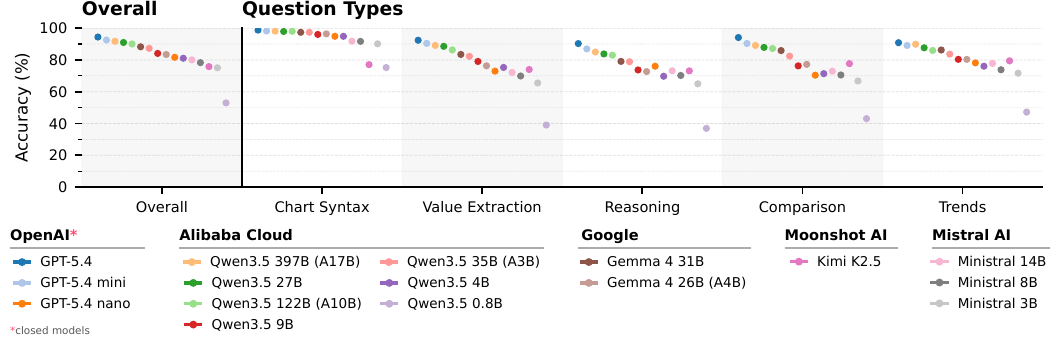}
  \caption{Evaluation accuracy across all 16 models. The left panel shows overall accuracy for each model, while the remaining panels show accuracy by question type.}
  \label{fig:evaluation}
\end{figure*}

\begin{figure}[t!]
  \centering
  \includegraphics{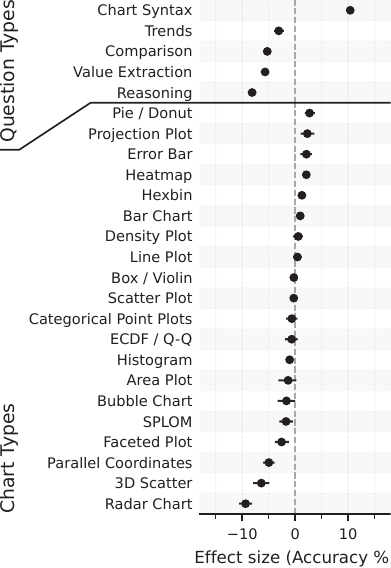}
  \caption{Question-type and chart-family effects after centering each group against each model's own overall accuracy.}
  \label{fig:evaluation-effects}
\end{figure}

To demonstrate one downstream use of the generated outputs, we evaluate 16 MLLMs on the chart-question pairs and analyze performance by question type and chart family. This evaluation is not intended as a definitive benchmark; rather, it serves as a diagnostic case study showing how aligned chart, code, and question-answer representations enable analyses beyond standard benchmark settings. By storing chart-family labels, dataset context, plotting code, and typed questions alongside each chart, the workflow enables diagnostic analyses that would be difficult with chart images and final answers alone.

We generated the corpus from public tabular datasets collected from the UCI machine learning repository~\cite{UCI}, using Qwen3.5-27B for all LLM-based generation and checking stages. We used spot checks of generated charts and question-answer pairs to identify recurring problems, including unreadable charts, ambiguous questions, and answers not supported by the rendered chart; these observations informed prompt revisions and filtering decisions before the downstream evaluation. The refinement filter discarded 725 of 2,228 chart candidates ($\sim 33\%$) whose severe readability or semantic-fit problems were not resolved within three correction attempts; the most common unresolved problems were low contrast, overplotting, unreadable axes, and unclear legends. The final collection contains 1,500 statistical charts derived from 74 datasets and 30,003 question-answer pairs, or approximately 20 questions per chart. The charts span 24 chart families, supporting analysis beyond common bar, line, and scatter plots. Each retained example stores the dataset description, generated chart, final plotting code, intermediate rendered versions, and refinement feedback, making individual examples traceable during inspection. This diversity is useful for diagnostic use cases because prior chart-understanding work has shown that model performance varies substantially across both task type and chart type \cite{Masry2022,Wang2025,Huang2025}.

For evaluation, each model receives the rendered chart, the dataset description, and one question, and returns a free-form answer without tool access. We then use Qwen3.5-9B as a separate LLM judge to compare the model response against the predefined answer and label it correct or incorrect, given the dataset description, question, reference answer, and candidate answer as context. For value-extraction questions, the judge accepts numerically close answers when the difference is consistent with approximate visual reading. We manually reviewed judge decisions during development and found them sufficiently consistent for aggregate diagnostic comparisons, while still treating the resulting accuracies as approximate rather than definitive correctness labels. We aggregate these judgments as accuracy and compute 95\% confidence intervals with a chart-level bootstrap, resampling whole charts to account for multiple questions attached to the same chart.

Figure~\ref{fig:evaluation} summarizes this illustrative diagnostic evaluation. The best overall model reaches about 94\% accuracy, while chart-syntax questions are nearly saturated for the strongest models, with peak accuracy over 99\%. In contrast, more visually and numerically grounded tasks remain more difficult: even the strongest models reach only 92\% accuracy on value extraction and 90\% on reasoning questions. This pattern is consistent with prior chart-understanding work, which similarly finds that direct chart reading is easier than questions requiring more complex visual, compositional, or numerical reasoning \cite{Masry2022,Hoque2022,Huang2025}.

Figure~\ref{fig:evaluation-effects} shows the same results as deviations relative to each model's overall accuracy. Chart-syntax questions are substantially easier than average, whereas comparison, value extraction, and reasoning questions are consistently harder. Chart family also matters: common chart families such as scatter plots, bar charts, heatmaps, and pie or donut charts tend to be easier, while radar charts, set and composition diagrams, parallel coordinates, and 3D scatter plots are more difficult. This matches prior work showing that performance varies across chart and task types \cite{Masry2022,Wang2025}, and reinforces the conclusion from EvoChart that the main bottleneck is not chart recognition alone, but accurate retrieval and reasoning under more complex visual conditions \cite{Huang2025}. Overall, the evaluation shows how chart-family labels and typed question-answer pairs expose systematic weaknesses obscured by aggregate accuracy alone.

\section{Conclusion}

We presented a structured LLM-based workflow for generating statistical charts from public tabular datasets through dataset screening, plot proposal, code generation, rendered-chart checking, refinement, description generation, and typed question-answer generation. The central contribution is the staged design: it turns chart generation into a sequence of inspectable steps rather than a single prompt-to-code call, and it stores the generated chart, plotting code, dataset context, description, question-answer pairs, and refinement feedback for later inspection and evaluation. More broadly, our results suggest that reliable chart generation requires treating rendering and visual validation as first-class steps, rather than relying solely on data- or code-level correctness.

The workflow nevertheless remains limited by its source data, models, prompts, and automatic checks. UCI datasets bias the generated corpus toward available public machine-learning tables. Refinement checks target visible readability and semantic-fit issues rather than guaranteeing full data or question correctness, and generated descriptions and question-answer pairs may still contain unsupported or ambiguous claims. At the same time, the workflow retains intermediate artifacts (code, dataset context, and refinement feedback), enabling manual inspection, filtering, and future human-in-the-loop validation. Finally, the LLM judge used in the downstream evaluation provides approximate aggregate labels rather than definitive correctness judgments. Future work should strengthen validation, characterize failure modes more systematically, and extend the workflow toward human-in-the-loop review and downstream chart-reasoning studies.

\subsection{Availability}
The workflow implementation is available at \url{https://github.com/pavlin-policar/llm-chart-generation} and licensed under the BSD 3-Clause license. An interactive viewer for the generated dataset is available at \url{https://llm-chart-generation.streamlit.app}.

\section*{Acknowledgments}
This work is supported by the Slovenian Research and Innovation Agency grants L7-70273 and P2-0209.

\section*{Author contributions}

\textbf{P.G.P.}: Conceptualization, Methodology, Formal analysis, Writing-Original draft, \textbf{A.P.}: Methodology, Investigation, Software, Writing-Review \& Editing, and \textbf{B.Z.}: Conceptualization, Supervision, Writing-Review \& Editing.

\bibliographystyle{abbrv}

\bibliography{main}

\end{document}